\documentclass[conference,a4paper]{IEEEtran}
\IEEEoverridecommandlockouts

\usepackage[hidelinks]{hyperref}
\usepackage[cmex10]{amsmath}
\usepackage{amssymb,amsfonts}
\usepackage{dblfloatfix}

\usepackage[ruled,vlined]{algorithm2e}
\usepackage{graphicx}
\graphicspath{{Figures/PDF/}{Figures/PNG/}}

\usepackage{booktabs}
\usepackage{siunitx}
\usepackage[numbers,compress]{natbib}
\usepackage{texnames}
\usepackage{bm,bbm}
\usepackage{orcidlink}
\usepackage{igarss}
\usepackage{listings}

\begin{document}

\title{

\uppercase{TerraQ: Spatiotemporal Question-Answering on Satellite Image Archives}

\thanks{	This work was supported by the ESA project DA4DTE (subcontract 202320239), the Horizon 2020 projects AI4Copernicus (GA No. 101016798) and STELAR (Grant No. 101070122), the Horizon Europe project FAIR2Adapt (GA no. 101188256), and the "NIK. D. XRYSOVERGI" scholarship of the Greek State Scholarship Foundation.}

}

\author{
    \IEEEauthorblockN{Sergios-Anestis Kefalidis\orcidlink{0009-0007-7648-9737}\textsuperscript{1}, 
    Konstantinos Plas\orcidlink{0009-0003-7613-4072}\textsuperscript{1,2}, 
    Manolis Koubarakis\orcidlink{0000-0002-1954-8338}\textsuperscript{1,2}}
    \IEEEauthorblockA{\textsuperscript{1}\textit{Dept. of Informatics and Telecommunications}, 
    \textit{National and Kapodistrian University of Athens}, Athens, Greece\\
    skefalidis@di.uoa.gr, kplas@di.uoa.gr, koubarak@di.uoa.gr\\
    \textsuperscript{2}\textit{Archimedes/Athena RC}, Marousi, Greece
    }
}

\maketitle
\begin{abstract}
	\EngineName{} is a spatiotemporal question-answering engine for satellite image archives. It is a natural language processing system that is built to process requests for satellite images satisfying certain criteria. The requests can refer to image metadata and entities from a specialized knowledge base (e.g., the Emilia-Romagna region). With it, users can make requests like “Give me a hundred images of rivers near ports in France, with less than 20\% snow coverage and more than 10\% cloud coverage”, thus making Earth Observation data more easily accessible, in-line with the current landscape of digital assistants.
\end{abstract}


\section{Introduction}

The field of Natural Language Processing is undergoing major advancements caused by the rapid development of Language Models~\cite{openai2024gpt4technicalreport, llama2, mistral}. An outcome of this development is the proliferation of digital assistants and natural language interfaces for all manners of systems and knowledge repositories (e.g., Alexa from Amazon, Siri from Apple, ChatGPT from OpenAI, and Claude from Anthropic). The resulting increase in accessibility enables non-technical users to intuitively interact with computer systems and access high-quality information, while also improving efficiency for expert users.  In this climate, the task of Knowledge-Graph Question-Answering (QA), also known as Text-to-SPARQL, is as relevant as ever~\cite{qa-survey1, qa-survey2}. 

QA systems take as input queries in natural language and generate semantically equivalent SPARQL\footnote{\url{https://www.w3.org/TR/sparql11-query/}} queries over a specific knowledge graph (KG). These SPARQL queries are subsequently executed on an RDF store, which in turn returns the answer. The answer can either be directly presented to the user, or integrated into a larger system and used as part of a Retrieval-Augmented Generation~\cite{rag} pipeline, as is the case with digital assistants that rely on knowledge grounding.

In our work, we are developing \EngineName{} a spatiotemporal QA engine for satellite image archives. 
User requests can refer to image metadata and geographic entities (e.g., the Loch Ness Lake or the city of Munich) both of which are included in the target knowledge graph. For example, ``Show me images of Athens with VV polarization.\%''. The goal of our research is to make Earth Observation data archives accessible via natural language, to the benefit of both novice and expert users.

\EngineName{} belongs in the same family of engines as GeoQA2~\cite{geoqa2} and EarthQA~\cite{earthqa}, two template-based question-answering engines previously developed by our group. GeoQA2 is a geospatial QA engine, and EarthQA is a satellite-image archive QA engine that reuses some components of GeoQA2 while also adding some additional specialized components. In comparison to these two systems, \EngineName{} has a number of advantages. First, it does away with template-based query generation. 
As a result, it is able to answer a wider array of questions and has improved accuracy. Second, \EngineName{} targets a purpose-built KG with high-quality geospatial information, allowing for more fine-grained results, which was one of the limitations of the original EarthQA paper. Third, unlike EarthQA, the engine does not use specialized components. All thematic information can be integrated into the core engine architecture, making the engine easier to adapt to different domains.

In this paper, we make the following original contributions:
\begin{enumerate}
    \item We present a specialized knowledge graph that interlinks geospatial information about natural features and administrative divisions with satellite image metadata. Knowledge graph resources are integrated into a hierarchical structure, making it easier to expand with additional geospatial information or thematic knowledge.
    \item We develop \EngineName{}, a spatiotemporal QA engine for image archives. The engine is able to answer simple and complex queries both reliably and quickly in dynamic fashion, while also avoiding the use of query-templates or computationally demanding neural models. 
\end{enumerate}

The version of \EngineName{} presented in this paper has been developed in the context of the European Space Agency project \textit{DA4DTE: Demonstrator Precursor Digital Assistant Interface for Digital Twin Earth\footnote{\url{http://da4dte.e-geos.earth/}}} and a demo is available publicly at {\url{http://terraq.di.uoa.gr/}}.


\section{Knowledge Graph}
\label{sec:kg}

To provide \EngineName{} with a powerful geospatial knowledge base, we compiled information from various sources and combined them under a common KG. Our aim was to create a polymorphic database of spatial resources, by integrating natural features and administrative geo-entities under a compact, non-complex ontology, that better facilitates the task of geospatial question answering.  The aforementioned facts were collected from the following sources:
\begin{itemize}
    \item GADM: We collected geospatial features from the Global Administrative Areas (\url{https://gadm.org/}) dataset, focusing on features located in Europe. Utilized features from this dataset include countries, cities, regional units, and national administrative divisions.
    \item Rivers, Points of Interest, and Ports: This dataset is provided by our partners in the DA4DTE project e-GEOS (\url{https://www.e-geos.it/}) and includes spatial characteristics for various features within these categories. In addition to spatial data, the dataset also contains comprehensive metadata for each feature.
    \item Sentinel-1 Images: We incorporated Sentinel-1 satellite image data, including metadata about the images and the satellite's location at the time each image was captured. Links to the images are stored in the knowledge graph, ensuring that they are easily accessible by the Question-Answering engine and external sources. Sentinel-1 images were collected for the years 2020 and 2021.
    \item Sentinel-2 Images: Similarly to the Sentinel-1 satellite images, we included image links and metadata from Sentinel-2 image collections to enhance the information available in our data model. Sentinel-2 images were collected for the years 2020, 2021 and 2022.
    \item Sea sectors: A collection of sea sectors covering global water spaces and oceans A total of 101 sea sectors along with their polygons were integrated into the knowledge graph of the Marine Regions~\cite{marine-regions} data source.
\end{itemize}

\textbf{Ontology.} We built our ontology on top of well-known and standardized ontologies, namely the YAGO2geo~\cite{DBLP:conf/semweb/KaralisMK19}
ontology and the GeoSPARQL~\cite{perry2012ogc} ontology. The main class of the knowledge graph is named Feature. The Feature class is extended by various subclasses that represent the knowledge provided by the various datasets that we examined in the previously, namely, rivers,
ports, pois (points of interest), Sentinel-1 and Sentinel-2 images and GADM geoentities. 


\textbf{Translation of named location labels to English.} While integrating our data, we noticed that many
geospatial features were named only in their original languages, with no English equivalents provided. For
instance, the city of Rome was listed only as "Roma" in Italian within the metadata. To provide a consistent framework and to simplify the task of recognizing labels for our Question-Answering engine, we implemented a comprehensive translation pipeline. Using the Mistral-7b LLM~\cite{mistral}, we were able to correctly identify and translate a total of 56657 labels from various European languages.
\section{The \EngineName{} Engine}
\label{sec:engine}

\EngineName{} 
consists of a number of components, each of which performs a specific task. Information is propagated from one component to the next. This pipeline is split in four distinct conceptual steps.


First, the WHERE clause is generated by combining basic SPARQL/GeoSPARQL building blocks. This is subsequently passed as additional input to the components responsible for the generation of the SELECT/ASK clause. When both clauses have been constructed, the query generator merges them and makes any necessary additions to construct a complete SPARQL query. In the last step, the query is rewritten to make use of materialized geospatial relations.

\begin{figure*}[h]
\begin{center}
\includegraphics[width=15cm]{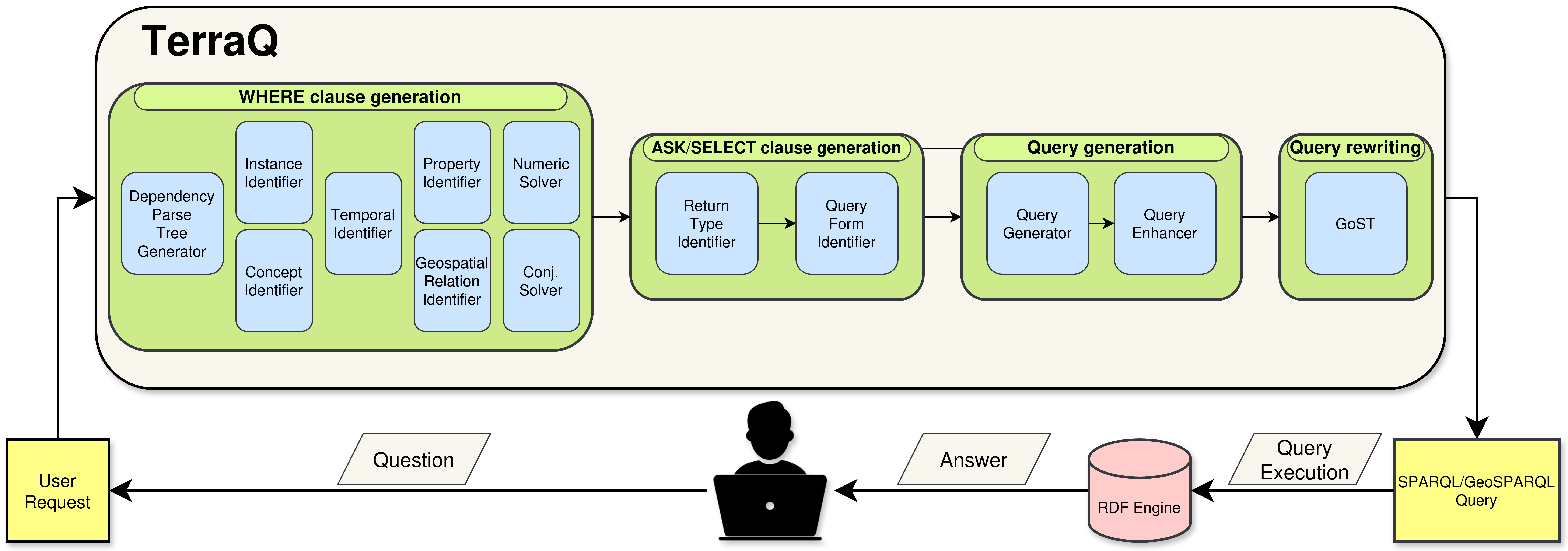}
\end{center}
\caption{The conceptual architecture of the \EngineName{} system}
\label{fig:system_arch}
\end{figure*}

The complete architecture of \EngineName{} is shown in Figure~\ref{fig:system_arch}. Below, we present the functionality of the system in detail. As our running example we use the request ``Show me all images taken in January 2021 with rivers less than 2km away from towns and forests in the Emilia Romagna region, having cloud coverage less than 10\%''.

\textbf{Dependency Parse Tree Generator.} This module generates a dependency parse tree of the input question using StanfordCoreNLP~\cite{corenlp}. The dependency parse tree is used to identify and store information.

\textbf{Instance Identifier.} This module does named-entity recognition and disambiguation.
In the example question, it identifies the entity “Emilia Romagna” and maps it to the resource \textit{yago:Emilia\_(region\_of\_Italy)} in the KG. The mapping to the KG resource happens in two steps. First, WAT~\cite{WAT} links the named entity to a Wikipedia page. Subsequently, the component searches the Knowledge Graph for the resource that best matches the entity returned by WAT.
In addition to identifying the instance, this component is responsible for creating the block that will be used in the WHERE clause for the identified instance. The generated block is the following:

\begin{lstlisting}[language=SPARQL]
<URI> geo:hasGeometry/geo:asWKT ?iWKTID .
\end{lstlisting}


\textbf{Concept Identifier.} This module identifies and maps concepts present in the input question to the appropriate resource of the KG ontology. For instance, from the example question, it will identify and map the concepts \textit{River, Town, Forest}. The mapping is done using a class label dictionary and string similarity based on n-grams.
Additionally, this component is responsible for creating the block that will be used in the WHERE clause for the identified concepts:

\begin{lstlisting}[language=SPARQL]
?cID a <URI> ; 
    geo:hasGeometry/geo:asWKT ?cWKTID .
\end{lstlisting}

At the end of the concept identification stage, and after all Instances have been identified, we employ a heuristic of consolidation between concepts and instances. Concepts and Instances that are not separated by any token are consolidated to reduce the complexity of the generated WHERE-clause and help the query generator produce a correct query. For example, in the question “Where is the Tagus river located?” only the Instance of Tagus is kept and the river concept is consolidated into it.

\textbf{Property Identifier.} The property identifier identifies attributes of features or types of features specified by the user in input questions and maps them to the corresponding properties in the knowledge graph. In the example question, the property ``cloud coverage'' of the type of feature image will be identified and mapped to the corresponding property in the KG.
For each identified concept, we try to match, using string similarity on n-grams, its properties to the words in the sentence. Matched properties are identified as candidate properties for this concept. Multiple concepts might have the same candidate property. To resolve this conflict, we introduced a heuristic that selects the closest concepts as the targets to the properties. This process is similar for instances inside the question.

Again, this component is also responsible for generating the block that will be used in the WHERE clause for the identified properties:

\begin{lstlisting}[language=SPARQL]
INSTANCE/CONCEPT_VARIABLE <URI> ?pID.
\end{lstlisting}

In addition, this component uses the dependency parse tree and Part-of-Speech tags to identify words that denote the use of comparatives and superlatives. These are subsequently matched to the appropriate Concept or Property, using a node-distance heuristic on the dependency parse tree.

\textbf{Spatial relation Identifier.} This module identifies spatial relations present in the input question and maps them to appropriate stSPARQL/GeoSPARQL functions. For instance, in the example question, it will identify the spatial relations ``in'' and ``away from'' and map them to \textit{geof:sfWithin} and \textit{geof:distance} respectively. Then these relations are mapped to the appropriate previously identified Instances and Concepts by using the following heuristic: 

\[
\text{distance} = \text{dependency\_parse\_tree\_distance} + \left(\frac{\text{word\_distance}}{100}\right)
\]

Again, this component is also responsible for generating the block that will be used in the WHERE clause:

\begin{lstlisting}[language=SPARQL]
FILTER (<URI> (FIRST_FEATURE, 
    SECOND_FEATURE))
\end{lstlisting}
and
\begin{lstlisting}[language=SPARQL]
FILTER (geof:distance (FIRST_FEATURE, 
    SECOND_FEATURE, uom:metre) 
    {<, >, <=, >=, =, ~} DISTANCE)
\end{lstlisting}

\textbf{Numeric Solver.} This module is responsible for identifying numbers, understanding their use in the input question and enhancing the previously identified elements with additional information. For this purpose we utilize Part-of-Speech tags and the previously described distance heuristic.

In our working example, ``less than 2km'' is matched to the spatial function of distance and ``less than 10\%'' is matched to the cloud coverage property.

\textbf{Conjunction Solver.} The Conjunction Solver is responsible for handling conjunctions, as those are identified by the dependency parse tree. To that end, it selects all edges of the parse tree tagged as ``conj:and''. The vertices connected by each of those edges are checked for meaningful conjunctions. Number-to-Property, Number-to-Number and Geospatial-to-Geospatial conjunctions are supported.
For Geospatial-to-Geospatial conjunctions additional spatial relations are generated and stored, as if they were created by the Geospatial Relation Identifier, according to the information provided by the vertices. In our example, this is the case with ``towns and forests''. Number and Property conjunctions faction similarly.

\textbf{Temporal Identifier.} This module uses HeidelTime~\cite{heideltime} to identify temporal keywords in the input question and annotates them with the appropriate date and/or duration. For instance, in the example input question it will identify ``January 2021'' and map it to 2021-01. 


\textbf{Return Type Identifier.} This module is responsible for identifying the expected form/type of the answer to the question. The supported types are \textit{ Name, Coordinates, Number-Property, Number-Count, Image }. For our example, \textit{Image} is the most appropriate return type.
For identifying the expected return types, this component leverages the sophisticated language understanding of Llama 2~\cite{llama2}. We fine-tune our model to output correctly formatted answers. A fallback mechanism that uses heuristics is provided to enable using \EngineName{} without hardware acceleration (GPU).

\textbf{Query Form Identifier.} This component is responsible for generating the final ASK/SELECT clause, which will be used by the query generator. It takes as input the return types generated by the return type identifier. For each expected return type, we follow an iterative approach as follows: If the type is \textit{Name}, we search for the next concept. If the type is \textit{Coordinates}, we seek the next concept or instance. When the return type is \textit{Number-Property}, we look for the next property, and if it's \textit{Number-Count}, we search for the next concept. Additionally, we enhance the query by introducing a COUNT aggregation and the necessary GROUP BY clauses. In the case of \textit{Image}, we insert the appropriate code in the query. To determine the 'next' object, we traverse the dependency parse tree.

\textbf{Query Generator.} The query generator is responsible for generating the final query. Within this stage of the pipeline, it assimilates all the information provided by the preceding components and combines them into a suitable, executable SPARQL or GeoSPARQL query. Information about superlatives, limits and other structures is taken into account in the generation process.

\textbf{Query Enhancer.} The query enhancer is an optional component responsible for modifying the query produced by the Query Generator to fix any mistakes and/or oversights. It is implemented using the Mistral-7b LLM fine-tuned on the dataset GeoQuestions1089~\cite{geoquestions1089}. It serves as a performance-enhancement module that increases the capacity of \EngineName{} to answer complex questions following the Execution Refinement paradigm~\cite{neural-interfaces}.

\textbf{GoST.} The GoST transpiler~\cite{geoquestions1089} takes the query generated by the Query Generator and rewrites it to use materialized geospatial relations if that is possible. Because geospatial relations like \textit{geof:sfWithin} are computationally expensive we do offline materialization using the tool JedAI-Spatial~\cite{jedai-spatial}.
\section{Evaluation}
\label{sec:eval}

To the best of our knowledge, there is no publically available dataset that is suitable for evaluating systems on the task of Text-to-SPARQL for Earth Observation archives. For this reason, we decided to deploy our engine as a pure geospatial QA engine and run an evaluation on the geospatial QA dataset GeoQuestions1089~\cite{geoquestions1089}. Although this did require some tinkering, since GeoQuestions1089 targets the YAGO2geo ontology, the process was straightforward and painless, and we believe that the resulting evaluation is useful for measuring the performance of most dimensions of our engine. Unfortunately, the questions in GeoQuestions1089 do not include temporal information.

To accept an answer as correct, it must match the gold result (included in GeoQuestions1089) exactly. We do not consider partially correct answers as correct. Likewise, for supersets of the answers in the gold set.

The results of our evaluation can be seen in Table~\ref{tab:geoq1089}. We benchmark \EngineName{} with the Query Enhancer disabled, since the model was fine-tuned on the same dataset, which would skew the results. We also compare our engine to GeoQA2 and the engine of Hamzei et al~\cite{hamzei}.


\begin{table}[hbt]
    \centering
    \caption{Evaluation on GeoQuestions1089$_c$ v1.1}\label{tab:geoq1089}
    \resizebox{\columnwidth}{!}{%
        \begin{tabular}{c c c c }
            \toprule
            \textbf{Category} & \textbf{GeoQA2 Accuracy} & \textbf{Hamzei Accuracy} & \textbf{\EngineName{} Accuracy} \\ \hline
            A & 53.52\% & 28.16\% & 60.56\% \\ \hline
            B & 62.68\% & 55.22\% & 73.13\% \\ \hline
            C & 48.36\% & 30.06\% & 47.71\% \\ \hline
            D & 9.09\% & 4.54\% & 22.73\% \\ \hline
            E & 24.81\% & 6.56\% & 23.36\% \\ \hline
            F & 28.57\% & 14.28\% & 38.10\% \\ \hline
            G & 36.30\% & 12.32\% & 28.77\% \\ \hline
            H & 23.07\% & 33.33\% & 40.17\% \\ \hline
            I & 21.73\% &8.69\% & 26.00\% \\ \hline
            ALL & 40.33\% & 25.92\% & 44.36\% \\
            \bottomrule
        \end{tabular}%
    }
\end{table}

We can see that \EngineName{} outperforms the previous state of the art in most question categories without utilizing templates. All in all, there is 4\% uplift in performance over the entire dataset, which is translated to a 10\% improvement relative to GeoQA2. The categories with performance regressions are caused by \EngineName{}’s more dynamic nature. \EngineName{} does not use predefined query templates and employs heuristics for a number of processes as previously described, these heuristics are not performing as well for those particular question categories. 
\section{Conclusion and Future Work}
\label{sec:conclusion}

The success of modern digital assistants has clearly shown that natural language interfaces for computer systems and knowledge repositories can be a great boon for productivity and accessibility. Users nowadays expect to be able to interact with their computers through natural language, reducing the barrier of technical knowledge required for utilizing modern computing capabilities.

This paper presents \EngineName{}, a spatiotemporal QA system for satellite image archives. Our engine targets a high-quality, purpose-built knowledge graph that contains Sentinel-1 and Sentinel-2 image metadata, as well as geospatial information for administrative divisions and natural features. Requests made in natural language are translated to SPARQL queries, which are subsequently executed by an RDF store. This enables users to request, in natural language, satellite images satisfying a number of complex spatial, temporal and thematic criteria. 

Our engine is easily deployable and responsive on commodity hardware, since it does not rely on exceedingly large LLMs. Instead, it utilizes a combination of small-scale LLMs, heuristics and expert knowledge. This development is another step towards our vision of making Earth Observation archives more accessible by both novice and expert users, no matter the available computing capacity.

In the future, we are planning on expanding the capabilities of our system by integrating Visual Question-Answering systems into our Knowledge Graph creation pipeline. This will enable users to express even more specific criteria for image selection, while also maintaining performance.

\small
\bibliographystyle{IEEEtranN}
\bibliography{references}

\begin{thebibliography}{18}
\providecommand{\natexlab}[1]{#1}
\providecommand{\url}[1]{#1}
\csname url@samestyle\endcsname
\providecommand{\newblock}{\relax}
\providecommand{\bibinfo}[2]{#2}
\providecommand{\BIBentrySTDinterwordspacing}{\spaceskip=0pt\relax}
\providecommand{\BIBentryALTinterwordstretchfactor}{4}
\providecommand{\BIBentryALTinterwordspacing}{\spaceskip=\fontdimen2\font plus
\BIBentryALTinterwordstretchfactor\fontdimen3\font minus \fontdimen4\font\relax}
\providecommand{\BIBforeignlanguage}[2]{{%
\expandafter\ifx\csname l@#1\endcsname\relax
\typeout{** WARNING: IEEEtranN.bst: No hyphenation pattern has been}%
\typeout{** loaded for the language `#1'. Using the pattern for}%
\typeout{** the default language instead.}%
\else
\language=\csname l@#1\endcsname
\fi
#2}}
\providecommand{\BIBdecl}{\relax}
\BIBdecl

\bibitem[OpenAI(2024)]{openai2024gpt4technicalreport}
\BIBentryALTinterwordspacing
OpenAI, ``Gpt-4 technical report,'' 2024. [Online]. Available: \url{https://arxiv.org/abs/2303.08774}
\BIBentrySTDinterwordspacing

\bibitem[Touvron et~al.(2023)Touvron, Martin, Stone, Albert, Almahairi, Babaei, Bashlykov, Batra, Bhargava, Bhosale, Bikel, Blecher, Canton{-}Ferrer, Chen, Cucurull, Esiobu, Fernandes, Fu, Fu, Fuller, Gao, Goswami, Goyal, Hartshorn, Hosseini, Hou, Inan, Kardas, Kerkez, Khabsa, Kloumann, Korenev, Koura, Lachaux, Lavril, Lee, Liskovich, Lu, Mao, Martinet, Mihaylov, Mishra, Molybog, Nie, Poulton, Reizenstein, Rungta, Saladi, Schelten, Silva, Smith, Subramanian, Tan, Tang, Taylor, Williams, Kuan, Xu, Yan, Zarov, Zhang, Fan, Kambadur, Narang, Rodriguez, Stojnic, Edunov, and Scialom]{llama2}
\BIBentryALTinterwordspacing
H.~Touvron, L.~Martin, K.~Stone, P.~Albert, A.~Almahairi, Y.~Babaei, N.~Bashlykov, S.~Batra, P.~Bhargava, S.~Bhosale, D.~Bikel, L.~Blecher, C.~Canton{-}Ferrer, M.~Chen, G.~Cucurull, D.~Esiobu, J.~Fernandes, J.~Fu, W.~Fu, B.~Fuller, C.~Gao, V.~Goswami, N.~Goyal, A.~Hartshorn, S.~Hosseini, R.~Hou, H.~Inan, M.~Kardas, V.~Kerkez, M.~Khabsa, I.~Kloumann, A.~Korenev, P.~S. Koura, M.~Lachaux, T.~Lavril, J.~Lee, D.~Liskovich, Y.~Lu, Y.~Mao, X.~Martinet, T.~Mihaylov, P.~Mishra, I.~Molybog, Y.~Nie, A.~Poulton, J.~Reizenstein, R.~Rungta, K.~Saladi, A.~Schelten, R.~Silva, E.~M. Smith, R.~Subramanian, X.~E. Tan, B.~Tang, R.~Taylor, A.~Williams, J.~X. Kuan, P.~Xu, Z.~Yan, I.~Zarov, Y.~Zhang, A.~Fan, M.~Kambadur, S.~Narang, A.~Rodriguez, R.~Stojnic, S.~Edunov, and T.~Scialom, ``Llama 2: Open foundation and fine-tuned chat models,'' \emph{CoRR}, vol. abs/2307.09288, 2023. [Online]. Available: \url{https://doi.org/10.48550/arXiv.2307.09288}
\BIBentrySTDinterwordspacing

\bibitem[Jiang et~al.(2023)Jiang, Sablayrolles, Mensch, Bamford, Chaplot, de~Las~Casas, Bressand, Lengyel, Lample, Saulnier, Lavaud, Lachaux, Stock, Scao, Lavril, Wang, Lacroix, and Sayed]{mistral}
\BIBentryALTinterwordspacing
A.~Q. Jiang, A.~Sablayrolles, A.~Mensch, C.~Bamford, D.~S. Chaplot, D.~de~Las~Casas, F.~Bressand, G.~Lengyel, G.~Lample, L.~Saulnier, L.~R. Lavaud, M.~Lachaux, P.~Stock, T.~L. Scao, T.~Lavril, T.~Wang, T.~Lacroix, and W.~E. Sayed, ``Mistral 7b,'' \emph{CoRR}, vol. abs/2310.06825, 2023. [Online]. Available: \url{https://doi.org/10.48550/arXiv.2310.06825}
\BIBentrySTDinterwordspacing

\bibitem[Diefenbach et~al.(2018)Diefenbach, L{\'{o}}pez, Singh, and Maret]{qa-survey1}
\BIBentryALTinterwordspacing
D.~Diefenbach, V.~L{\'{o}}pez, K.~D. Singh, and P.~Maret, ``Core techniques of question answering systems over knowledge bases: a survey,'' \emph{Knowl. Inf. Syst.}, vol.~55, no.~3, pp. 529--569, 2018. [Online]. Available: \url{https://doi.org/10.1007/s10115-017-1100-y}
\BIBentrySTDinterwordspacing

\bibitem[H{\"{o}}ffner et~al.(2017)H{\"{o}}ffner, Walter, Marx, Usbeck, Lehmann, and Ngomo]{qa-survey2}
\BIBentryALTinterwordspacing
K.~H{\"{o}}ffner, S.~Walter, E.~Marx, R.~Usbeck, J.~Lehmann, and A.~N. Ngomo, ``Survey on challenges of question answering in the semantic web,'' \emph{Semantic Web}, vol.~8, no.~6, pp. 895--920, 2017. [Online]. Available: \url{https://doi.org/10.3233/SW-160247}
\BIBentrySTDinterwordspacing

\bibitem[Lewis et~al.(2020)Lewis, Perez, Piktus, Petroni, Karpukhin, Goyal, K{\"{u}}ttler, Lewis, Yih, Rockt{\"{a}}schel, Riedel, and Kiela]{rag}
\BIBentryALTinterwordspacing
P.~S.~H. Lewis, E.~Perez, A.~Piktus, F.~Petroni, V.~Karpukhin, N.~Goyal, H.~K{\"{u}}ttler, M.~Lewis, W.~Yih, T.~Rockt{\"{a}}schel, S.~Riedel, and D.~Kiela, ``Retrieval-augmented generation for knowledge-intensive {NLP} tasks,'' in \emph{Advances in Neural Information Processing Systems 33: Annual Conference on Neural Information Processing Systems 2020, NeurIPS 2020, December 6-12, 2020, virtual}, H.~Larochelle, M.~Ranzato, R.~Hadsell, M.~Balcan, and H.~Lin, Eds., 2020. [Online]. Available: \url{https://proceedings.neurips.cc/paper/2020/hash/6b493230205f780e1bc26945df7481e5-Abstract.html}
\BIBentrySTDinterwordspacing

\bibitem[Kefalidis et~al.(2024)Kefalidis, Punjani, Tsalapati, Plas, Pollali, Maret, and Koubarakis]{geoqa2}
\BIBentryALTinterwordspacing
S.~Kefalidis, D.~Punjani, E.~Tsalapati, K.~Plas, M.~Pollali, P.~Maret, and M.~Koubarakis, ``The question answering system geoqa2 and a new benchmark for its evaluation,'' \emph{Int. J. Appl. Earth Obs. Geoinformation}, vol. 134, p. 104203, 2024. [Online]. Available: \url{https://doi.org/10.1016/j.jag.2024.104203}
\BIBentrySTDinterwordspacing

\bibitem[Punjani et~al.(2023)Punjani, Koubarakis, and Tsalapati]{earthqa}
\BIBentryALTinterwordspacing
D.~Punjani, M.~Koubarakis, and E.~Tsalapati, ``Earthqa: {A} question answering engine for earth observation data archives \({}^{\mbox{*}}\),'' in \emph{{IEEE} International Geoscience and Remote Sensing Symposium, {IGARSS} 2023, Pasadena, CA, USA, July 16-21, 2023}.\hskip 1em plus 0.5em minus 0.4em\relax {IEEE}, 2023, pp. 1396--1399. [Online]. Available: \url{https://doi.org/10.1109/IGARSS52108.2023.10282475}
\BIBentrySTDinterwordspacing

\bibitem[Institute(2021)]{marine-regions}
\BIBentryALTinterwordspacing
F.~M. Institute, ``Global oceans and seas, version 1,'' 2021. [Online]. Available: \url{https://doi.org/10.14284/542}
\BIBentrySTDinterwordspacing

\bibitem[Karalis et~al.(2019)Karalis, Mandilaras, and Koubarakis]{DBLP:conf/semweb/KaralisMK19}
\BIBentryALTinterwordspacing
N.~Karalis, G.~M. Mandilaras, and M.~Koubarakis, ``Extending the {YAGO2} knowledge graph with precise geospatial knowledge,'' in \emph{The Semantic Web - {ISWC} 2019 - 18th International Semantic Web Conference, Auckland, New Zealand, October 26-30, 2019, Proceedings, Part {II}}, ser. Lecture Notes in Computer Science, C.~Ghidini, O.~Hartig, M.~Maleshkova, V.~Sv{\'{a}}tek, I.~F. Cruz, A.~Hogan, J.~Song, M.~Lefran{\c{c}}ois, and F.~Gandon, Eds., vol. 11779.\hskip 1em plus 0.5em minus 0.4em\relax Springer, 2019, pp. 181--197. [Online]. Available: \url{https://doi.org/10.1007/978-3-030-30796-7\_12}
\BIBentrySTDinterwordspacing

\bibitem[{Matthew Perry} and {John Herring}(2012)]{perry2012ogc}
\BIBentryALTinterwordspacing
{Matthew Perry} and {John Herring}, ``{OGC} {GeoSPARQL} - {A} {Geographic} {Query} {Language} for {RDF} {Data},'' Open Geospatial Consortium, {OGC} {Implementation} {Standard} OGC 11-052r4, Sep. 2012. [Online]. Available: \url{http://www.opengis.net/doc/IS/geosparql/1.0}
\BIBentrySTDinterwordspacing

\bibitem[Manning et~al.(2014)Manning, Surdeanu, Bauer, Finkel, Bethard, and McClosky]{corenlp}
\BIBentryALTinterwordspacing
C.~D. Manning, M.~Surdeanu, J.~Bauer, J.~R. Finkel, S.~Bethard, and D.~McClosky, ``The stanford corenlp natural language processing toolkit,'' in \emph{Proceedings of the 52nd Annual Meeting of the Association for Computational Linguistics, {ACL} 2014, June 22-27, 2014, Baltimore, MD, USA, System Demonstrations}.\hskip 1em plus 0.5em minus 0.4em\relax The Association for Computer Linguistics, 2014, pp. 55--60. [Online]. Available: \url{https://doi.org/10.3115/v1/p14-5010}
\BIBentrySTDinterwordspacing

\bibitem[Piccinno and Ferragina(2014)]{WAT}
\BIBentryALTinterwordspacing
F.~Piccinno and P.~Ferragina, ``From tagme to {WAT:} a new entity annotator,'' in \emph{ERD'14, Proceedings of the First {ACM} International Workshop on Entity Recognition {\&} Disambiguation, July 11, 2014, Gold Coast, Queensland, Australia}, D.~Carmel, M.~Chang, E.~Gabrilovich, B.~P. Hsu, and K.~Wang, Eds.\hskip 1em plus 0.5em minus 0.4em\relax {ACM}, 2014, pp. 55--62. [Online]. Available: \url{https://doi.org/10.1145/2633211.2634350}
\BIBentrySTDinterwordspacing

\bibitem[Str{\"{o}}tgen and Gertz(2010)]{heideltime}
\BIBentryALTinterwordspacing
J.~Str{\"{o}}tgen and M.~Gertz, ``Heideltime: High quality rule-based extraction and normalization of temporal expressions,'' in \emph{Proceedings of the 5th International Workshop on Semantic Evaluation, SemEval@ACL 2010, Uppsala University, Uppsala, Sweden, July 15-16, 2010}, K.~Erk and C.~Strapparava, Eds.\hskip 1em plus 0.5em minus 0.4em\relax The Association for Computer Linguistics, 2010, pp. 321--324. [Online]. Available: \url{https://aclanthology.org/S10-1071/}
\BIBentrySTDinterwordspacing

\bibitem[Kefalidis et~al.(2023)Kefalidis, Punjani, Tsalapati, Plas, Pollali, Mitsios, Tsokanaridou, Koubarakis, and Maret]{geoquestions1089}
\BIBentryALTinterwordspacing
S.~Kefalidis, D.~Punjani, E.~Tsalapati, K.~Plas, M.~Pollali, M.~Mitsios, M.~Tsokanaridou, M.~Koubarakis, and P.~Maret, ``Benchmarking geospatial question answering engines using the dataset geoquestions1089,'' in \emph{The Semantic Web - {ISWC} 2023 - 22nd International Semantic Web Conference, Athens, Greece, November 6-10, 2023, Proceedings, Part {II}}, ser. Lecture Notes in Computer Science, T.~R. Payne, V.~Presutti, G.~Qi, M.~Poveda{-}Villal{\'{o}}n, G.~Stoilos, L.~Hollink, Z.~Kaoudi, G.~Cheng, and J.~Li, Eds., vol. 14266.\hskip 1em plus 0.5em minus 0.4em\relax Springer, 2023, pp. 266--284. [Online]. Available: \url{https://doi.org/10.1007/978-3-031-47243-5\_15}
\BIBentrySTDinterwordspacing

\bibitem[Hong et~al.(2024)Hong, Yuan, Zhang, Chen, Dong, Huang, and Huang]{neural-interfaces}
\BIBentryALTinterwordspacing
Z.~Hong, Z.~Yuan, Q.~Zhang, H.~Chen, J.~Dong, F.~Huang, and X.~Huang, ``Next-generation database interfaces: {A} survey of llm-based text-to-sql,'' \emph{CoRR}, vol. abs/2406.08426, 2024. [Online]. Available: \url{https://doi.org/10.48550/arXiv.2406.08426}
\BIBentrySTDinterwordspacing

\bibitem[Papamichalopoulos et~al.(2022)Papamichalopoulos, Papadakis, Mandilaras, Siampou, Mamoulis, and Koubarakis]{jedai-spatial}
\BIBentryALTinterwordspacing
M.~Papamichalopoulos, G.~Papadakis, G.~Mandilaras, M.~D. Siampou, N.~Mamoulis, and M.~Koubarakis, ``Three-dimensional geospatial interlinking with jedai-spatial,'' \emph{CoRR}, vol. abs/2205.01905, 2022. [Online]. Available: \url{https://doi.org/10.48550/arXiv.2205.01905}
\BIBentrySTDinterwordspacing

\bibitem[Hamzei et~al.(2022)Hamzei, Tomko, and Winter]{hamzei}
\BIBentryALTinterwordspacing
E.~Hamzei, M.~Tomko, and S.~Winter, ``Translating place-related questions to geosparql queries,'' in \emph{{WWW} '22: The {ACM} Web Conference 2022, Virtual Event, Lyon, France, April 25 - 29, 2022}, F.~Laforest, R.~Troncy, E.~Simperl, D.~Agarwal, A.~Gionis, I.~Herman, and L.~M{\'{e}}dini, Eds.\hskip 1em plus 0.5em minus 0.4em\relax {ACM}, 2022, pp. 902--911. [Online]. Available: \url{https://doi.org/10.1145/3485447.3511933}
\BIBentrySTDinterwordspacing

\end{thebibliography}

\end{document}